\documentclass[journal]{IEEEtran}
\usepackage{amsmath,amsfonts}
\usepackage{algorithmic}
\usepackage{algorithm}
\usepackage{array}
\usepackage[caption=false,font=normalsize,labelfont=sf,textfont=sf]{subfig}
\usepackage{textcomp}
\usepackage{stfloats}
\usepackage{url}
\usepackage{verbatim}
\usepackage{graphicx}
\usepackage{cite}
\usepackage[bookmarks,colorlinks,linkcolor=red, anchorcolor=blue, citecolor=blue]{hyperref}
\usepackage[abs]{overpic}
\usepackage{booktabs, multirow, multicol, makecell}
\RequirePackage{xspace}

\DeclareGraphicsExtensions{.pdf,.jpg,.png,.jpeg}

\makeatletter
\DeclareRobustCommand\onedot{\futurelet\@let@token\@onedot}
\def\@onedot{\ifx\@let@token.\else.\null\fi\xspace}

\def\eg{\emph{e.g}\onedot} 
\def\ie{\emph{i.e}\onedot}

\def\etal{\emph{et al}\onedot}
\makeatother

\newcommand{\tabincell}[2]{\begin{tabular}{@{}#1@{}}#2\end{tabular}}
\usepackage{xcolor}
\newcommand{\changes}{}

\begin{document}

\title{NIR-Assisted Image Denoising: A Selective Fusion Approach and A Real-World Benchmark Dataset}

\author{Rongjian Xu, Zhilu Zhang\thanks{* Corresponding author }\IEEEauthorrefmark{1}, Renlong Wu, and Wangmeng Zuo,~\IEEEmembership{Senior Member,~IEEE}
\thanks{R. Xu is with the Faculty of Computing, Harbin Institute of Technology, Harbin, China. (E-mail: ronjon.xu@gmail.com)}
\thanks{Z. Zhang is with the Faculty of Computing, Harbin Institute of Technology, Harbin, China. (E-mail: cszlzhang@outlook.com)}
\thanks{R. Wu is with the Faculty of Computing, Harbin Institute of Technology, Harbin, China. (E-mail:hirenlongwu@gmail.com)}
\thanks{W. Zuo is with the Faculty of Computing, Harbin Institute of Technology, Harbin, China. (E-mail: wmzuo@hit.edu.cn)}}

\maketitle

\begin{abstract}
Despite the significant progress in image denoising, it is still challenging to restore fine-scale details while removing noise, especially in extremely low-light environments. Leveraging near-infrared (NIR) images to assist visible RGB image denoising shows the potential to address this issue, becoming a promising technology. Nonetheless, existing works still struggle with taking advantage of NIR information effectively for real-world image denoising, due to the content inconsistency between NIR-RGB images and the scarcity of real-world paired datasets. To alleviate the problem, we propose an efficient Selective Fusion Module (SFM), which can be plug-and-played into the advanced denoising networks to merge the deep NIR-RGB features. Specifically, we sequentially perform the global and local modulation for NIR and RGB features, and then integrate the two modulated features. Furthermore, we present a Real-world NIR-Assisted Image Denoising (Real-NAID) dataset, which covers diverse scenarios as well as various noise levels. Extensive experiments on both synthetic and our real-world datasets demonstrate that the proposed method achieves better results than state-of-the-art ones. The dataset, codes, and pre-trained models are available at \url{https://github.com/ronjonxu/NAID}.
\end{abstract}


\section{Introduction}
\IEEEPARstart{I}{n} low-light conditions, it's common to use short exposure time and high ISO in imaging to prevent motion blur, while this setting inevitably introduces noise due to the limited number of photons captured by camera.
With the development of deep learning~\cite{he2016deep,liang2021swinir,vaswani2017attention,zha1,zha2,zha3}, many image denoising methods~\cite{zhang2017beyond,zhang2018ffdnet,abdelhamed2020ntire,zamir2022restormer,wang2022uformer,li2023spatially,xformer} have been proposed to remove the noise. 
Although great progress has been achieved, it is still challenging for these methods to recover fine-scale details faithfully due to the severely ill-posed nature of denoising.
An alternative solution is multi-frame denoising~\cite{mildenhall2018burst,godard2018deep,pearl2022nan,wu2023rbsr,zhang2022self,zhang2024bracketing}, in which multiple successive frames are merged to improve performance.
However, it is susceptible to the misalignment between frames, and may be less effective in facing dynamic scenes.

Fortunately, near-infrared (NIR) images with low noise can be captured at a cheap cost and utilized to enhance the denoising of visible RGB images~\cite{lv2020integrated,wu2020learn,jin2022darkvisionnet,wan2022purifying}, dubbed NIR-Assisted Image Denoising (NAID).
Specifically, on the one hand, the NIR band lies outside the range of the human visible spectrum.
It enables us to turn on NIR light that is imperceptible to humans, thus capturing NIR images~\cite{fredembach2008colouring} with a low noise level. 
On the other hand, modern CMOS sensors are sensitive to partial near-infrared wavelengths~\cite{xiong2021seeing}, thus allowing NIR signals to be acquired cheaply and conveniently.

\begin{figure}[t!]
    \centering
    \includegraphics[width=1\linewidth]{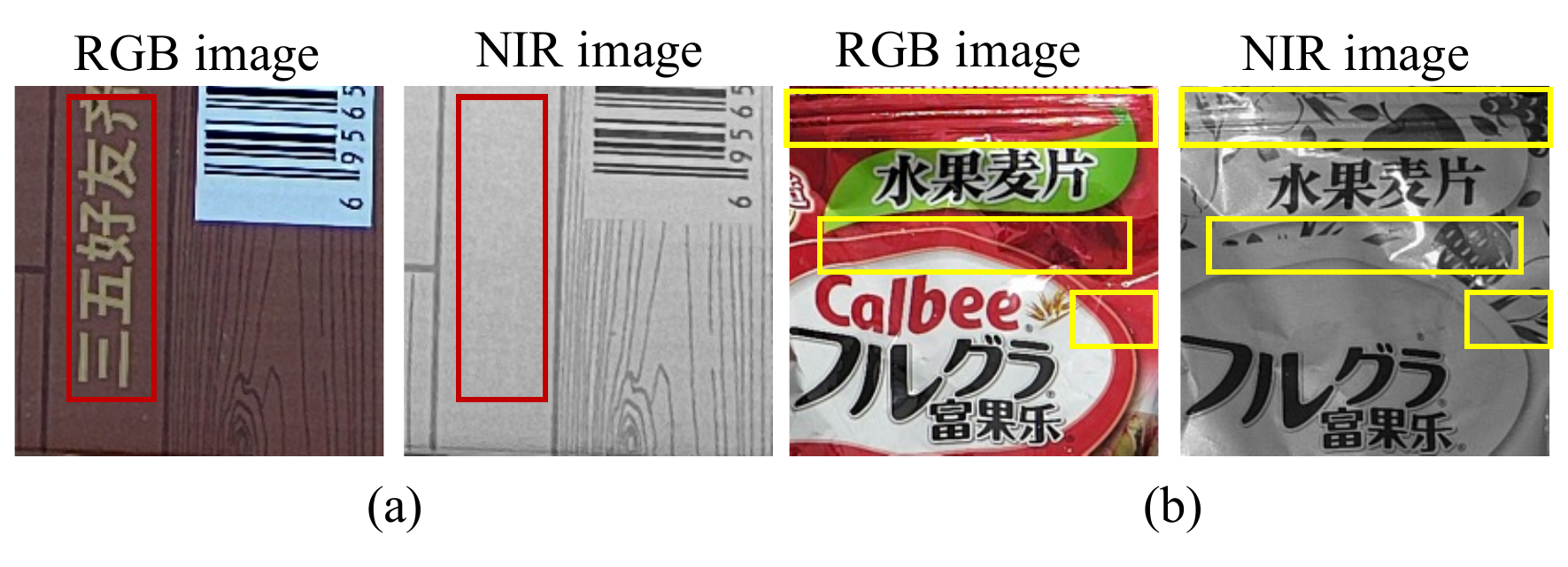}
    \vspace{-9mm}
    \caption{Examples of the structure discrepancy between NIR-RGB images. ($\mathbf{a}$) The structure is visible in the RGB image but not in the NIR image, as shown in the red box.
    ($\mathbf{b}$) The structure is visible in the NIR image but not in the RGB image, as shown in the yellow box.
 }
    \label{fig:structure_example}
    \vspace{0mm}
\end{figure}

\changes{Nevertheless, the color, brightness, and structure inconsistencies between NIR and RGB contents limit the positive effect of NIR images in denoising. 
First, NIR images are monochromatic, which leads to color discrepancy between the two modalities.
Second, NIR images are captured under additional NIR light, and objects exhibit varying spectral reflectance in the RGB (400-720 nm) and NIR band (720-1100 nm).
These lead to the brightness inconsistency.}
Thirdly, the NIR images may `less-see' or `more-see' the objects than the visible light ones, leading to structure inconsistency between the two modalities~\cite{fredembach2008colouring}.
\changes{On the one hand, different from visible RGB images, NIR images tend to provide information pertinent to material properties rather than the color of objects.
An example can be seen in the red box in Fig.~\ref{fig:structure_example} (a). 
As the yellow texts and the background have similar NIR spectral reflectance, the background and texts are indistinguishable in the NIR image while the RGB image clearly contains textual information.
On the other hand, NIR has stronger penetrating power than visible light, leading NIR images to `more-see' the objects.
As shown in the yellow box in Fig.~\ref{fig:structure_example} (b), the NIR image exhibits extra fruit patterns while these patterns are absent in the RGB image. }
DVD~\cite{jin2022darkvisionnet} and SANet~\cite{SANet} have noticed this problem, but their solutions are complex and less effective.
In this work, we focus on the inconsistency issue and aim to circumvent its adverse effects on combinations of NIR and RGB images in an efficient manner.
On the one hand, we introduce a Global Modulation Module (GMM) and a Local Modulation Module (LMM) to deal with color and structure inconsistency issues respectively. 
They predict and assign soft weights to NIR and RGB features, thus preparing for subsequent feature fusion. 
The combination of GMM, LMM, and fusion operation is called the Selective Fusion Module (SFM).
On the other hand, to take full advantage of advanced denoising network architectures~\cite{wang2022uformer,zamir2022restormer,NAFNet,xformer}, we expect SFM to be integrated into varying networks easily and efficiently. Thus, we design SFM according to the principles of simplicity and pluggability.

Additionally, due to the lack of real-world paired datasets, most existing methods can only process synthetic noisy images.
NIR-assisted real-world noise removal is rarely explored.
To address the issue, we construct a real-world NAID dataset, named Real-NAID.
It encompasses diverse scenarios and various noise levels, providing a valuable resource for evaluating and promoting research in this field.
Extensive experiments are conducted on both synthetic DVD~\cite{jin2022darkvisionnet} and our Real-NAID datasets.
The results show that the proposed method performs better than state-of-the-art ones.

Our contributions can be summarized as follows: 
(1) For NIR-assisted image denoising, we propose a plug-and-play selective fusion module to handle content inconsistency issues between NIR-RGB images, which assigns appropriate fusion weights to the deep NIR and RGB features by global and local modulation modules.

(2) We bring NIR-assisted image denoising to the real world by constructing a paired real-world dataset, which covers diverse scenarios and various noise levels. 

(3) Extensive experiments on both synthetic and our real-world datasets demonstrate that the proposed method achieves better than state-of-the-art ones quantitatively and qualitatively. 

\section{Relate Work}

\subsection{Single Image Denoising}
With the advancements in deep learning~\cite{ronneberger2015u,he2016deep,vaswani2017attention}, numerous single-image denoising methods~\cite{zhang2017beyond,zhang2018ffdnet,abdelhamed2020ntire,chen2021pre,zamir2022restormer,wang2022uformer,zhang2022self,li2023spatially} have emerged.
DnCNN~\cite{zhang2017beyond} pioneers the utilization of deep learning techniques and surpasses traditional patch-based methods~\cite{buades2005non, dabov2007image, gu2014weighted} on Gaussian noise removal.
Recently, some methods~\cite{zamir2021multi, wang2022uformer, zamir2022restormer,NAFNet} are developed with advanced architectures.
For example, MPRNet~\cite{zamir2021multi} applies a multi-stage architecture for progressive image restoration and achieves remarkable performance. 
Uformer~\cite{wang2022uformer} introduces the locally-enhanced transformer by employing the non-overlapping window-based self-attention. 
Restormer~\cite{zamir2022restormer} further reduces the computation cost by modifying the self-attention calculation from the spatial dimension to channel one.
NAFNet~\cite{NAFNet} proposes a simple baseline that does not apply nonlinear activation.
Despite the significant progress achieved by these methods, the performance is still unsatisfactory when handling images with high-level noise captured under low-light conditions, due to the severely ill-posed nature of denoising.

\subsection{NIR-assited Image Restoration}
Compared to single-image restoration, NIR images have the potential to assist in restoring details from degraded images.
The earlier work~\cite{10.1145/1531326.1531402} utilizes gradient constraints for NIR-assisted image denoising. 
Wang \etal~\cite{wang2019near} further improves the performance with deep learning methods.
SSN~\cite{wu2020learn} proposes a multi-task deep network with state synchronization modules.
TC-GAN~\cite{TC-GAN} fuses NIR images and RGB ones based on a texture conditional generative adversarial network.
DCMAN~\cite{DCMAN} employs spatial-temporal-spectral priors to introduce NIR videos for low-light RGB video restoration.
However, these methods have overlooked the color and structure inconsistency issues between the NIR images and RGB ones.
CCDFuse~\cite{zhao2023cddfuse} addresses this issue by combining the local modeling ability of convolutional blocks and the non-local modeling ability of transformer ones to extract local and global features of NIR and RGB images respectively.
SANet~\cite{SANet} proposes a guided denoising framework by estimating a clean structure map for the noisy RGB image.
Wan \etal~\cite{wan2022purifying} disentangle the color and structure components from the NIR images and RGB ones.
Besides, a few works~\cite{CUNet,MNNet,jin2022darkvisionnet} incorporate different priors into the network design, like sparse coding~\cite{CUNet}, deep implicit prior~\cite{MNNet} and deep inconsistency prior~\cite{jin2022darkvisionnet}.
However, their intricate designs make it difficult to integrate them into existing advanced restoration networks, hindering their extensions and improvements.

\subsection{Datasets for NIR-Assisted Image Restoration}
Existing NIR-RGB datasets suffer from limitations such as scarcity of data samples~\cite{10.1145/1531326.1531402}, absence of paired real-world RGB noisy images~\cite{brown2011multi, zhi2018deep, jin2022darkvisionnet, lv2020integrated}, or lack of public accessibility~\cite{stereodarkflash, lv2020integrated}. 
For instance, Krishnan \etal~\cite{10.1145/1531326.1531402} develop a prototype camera to capture image pairs under varying low-light conditions but only containing 5 image pairs.
Its size is too small to fulfill the demands of data-driven deep-learning algorithms.
IVRG~\cite{brown2011multi} and RGB-NIR Srereo~\cite{zhi2018deep} construct datasets consisting of RGB and NIR image pairs for image recognition and stereo matching, respectively.
DVD~\cite{jin2022darkvisionnet} captures images within a controlled light-box environment. 
However, these datasets only comprise clean RGB and NIR image pairs, lacking real-world noisy RGB images. 
Burst Dataset~\cite{stereodarkflash} captures real-world noisy images by a mobile imaging device that is sensitive to both near-infrared and near-ultraviolet signals. 
Lv \etal ~\cite{lv2020integrated} introduces the VIS-NIR-MIX dataset which utilizes a motorized rotator to manipulate illumination conditions.
But they are not publicly available.
In this work, we introduce a real-world NAID benchmark dataset, which encompasses diverse scenarios and various noise levels.

\section{Real-NAID: Real-World NIR-Assisted Image Denoising Dataset}
\label{section:RealNAID_dataset}

\begin{figure*}[t!]
    \centering
    \includegraphics[width=0.99\linewidth]{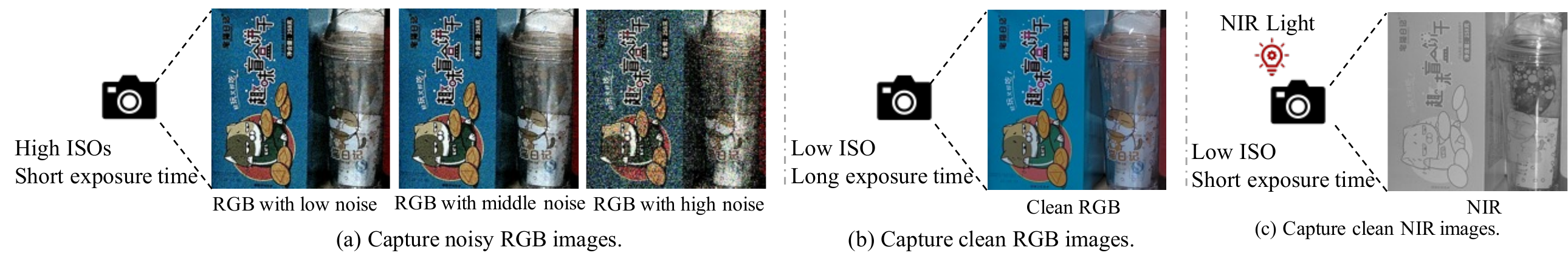}
    \vspace{-4mm}
    \caption{The construction of Real-NAID dataset. ($\mathbf{a}$) Capture noisy RGB images with three types of high ISO and short exposure time.
    ($\mathbf{b}$) Capture clean RGB images with low ISO and long exposure time. 
    ($\mathbf{c}$) Turn on the NIR light, then capture the clean NIR images with low ISO and short exposure time. }
    \label{fig:dataset_pipeline}
    \vspace{-1mm}
\end{figure*}

Existing publicly available NAID datasets generally lack real-world noisy RGB images paired with clean RGB and NIR images, which limits the investigation in real-world NAID.
\changes{
Compared with the synthetic data, the noise in the real world is more complex and more difficult to remove, which makes it more challenging to fuse NIR and RGB images.
Take synthetic DVD~\cite{jin2022darkvisionnet} dataset as an example, its noisy images are synthesized by simply adding Gaussian and Poisson noise to clean images, and the noise parameters are casually selected.
Actually, this simple noise model is far from real-world complex noise.
On the one hand, the noise parameters are Inaccurately calibrated.
On the other hand, some image noise, \eg, row noise~\cite{wei2020physics}, quantization noise~\cite{wei2020physics}, and fixed pattern noise~\cite{FPN}, are beyond the Gaussian and Poisson distribution. 
These noises are indispensable in low-light environments but are not considered.
}
To break such a limitation, we build a real-world NAID dataset, named Real-NAID.
Specifically, we employ high ISO and short exposure time to capture the real-world noisy RGB images, as shown in Fig.~\ref{fig:dataset_pipeline} (a).
For capturing the corresponding clean RGB images, we lower the ISO of the camera and appropriately increase the exposure time, as shown in Fig.~\ref{fig:dataset_pipeline} (b).
To obtain paired NIR images, we activate NIR light to ensure a sufficient supply of NIR illumination and then capture them with a dedicated NIR camera, as shown in Fig.~\ref{fig:dataset_pipeline} (c).
All images are captured with the Huawei X2381-VG surveillance camera, which is equipped with a built-in NIR illuminator specifically designed for capturing NIR images \changes{with spectral values between 750 nm and 1100 nm}.
To ensure image registration among multiple captures, we carefully position the camera and develop a remote control application to capture images of static objects. 
\changes{When capturing clean RGB images, the camera's ISO is set to 600, as it does not introduce significant noise while allowing us to set a short exposure time to avoid blurry content.}
When capturing RGB images with low, middle, and high noise levels, we set ISO to 4000, 12000, and 32000, respectively.
It is worth noting that we adjust the exposure time of each noise level to keep the brightness of noisy images relatively constant.

Besides, following DVD~\cite{jin2022darkvisionnet}, we crop all images from 2048 $\times$ 3840 resolutions into 2048 $\times$ 2160 to mitigate the vignetting effect.
In total, the dataset comprises 100 scenes with diverse contents, and each scene has three noisy images with various noise levels.
70 scenes are randomly sampled as the training set and the remaining 30 ones are used for the testing set.
In addition, we compare our Real-NAID dataset with other existing NIR-RGB datasets to demonstrate its strengths, as shown in Table~\ref{tab:dataset_table}.

\begin{table}[t!] 
    \setlength{\tabcolsep}{2pt}
    %
    \renewcommand\arraystretch{1.25}
    \small
    \begin{center}
	\caption{Comparisons of some existing datasets consisting of paired NIR and RGB images. `Public' refers to its public accessibility. `Number' denotes the number of paired images.}
        \vspace{-1mm}
	\label{tab:dataset_table}
        \scalebox{0.87}{
	\begin{tabular}{lcccc}
		\toprule
        Dataset & Real-Noise  & Public & Number & Resolution  \\
        \midrule
           RGB-NIR Video \cite{DCMAN} &   &  & 11444 & $ 1280 \times 720 $ \\
           RGB-NIR Stereo \cite{zhi2018deep} &     & \checkmark & 42000 &  $\sim 582 \times 492$  \\
            IVRG \cite{brown2011multi} &     & \checkmark & 477 &  $\sim 1024 \times 680$  \\
		DVD \cite{jin2022darkvisionnet} &    &  \checkmark & 307 & $1792 \times 1008 $  \\
            Burst Dataset \cite{stereodarkflash} & \checkmark  &  & 121 & $ 512 \times 512 $ \\ 
            VIS-NIR-MIX \cite{lv2020integrated} & \checkmark &  & 206 & $\sim 3072 \times 2048$  \\
            Dark Flash Photography \cite{10.1145/1531326.1531402} & \checkmark  & \checkmark & 5 & $\sim 1400 \times 1000$  \\
		Real-NAID (Ours) & \checkmark  & \checkmark & 300 & $ 2160 \times 2048 $  \\
		\bottomrule
	\end{tabular}}
    \end{center}
    \vspace{-2mm}
\end{table}

\section{Method}
\subsection{Problem Formation}
\begin{figure*}[t!]
    \centering
    \includegraphics[width=0.95\linewidth]{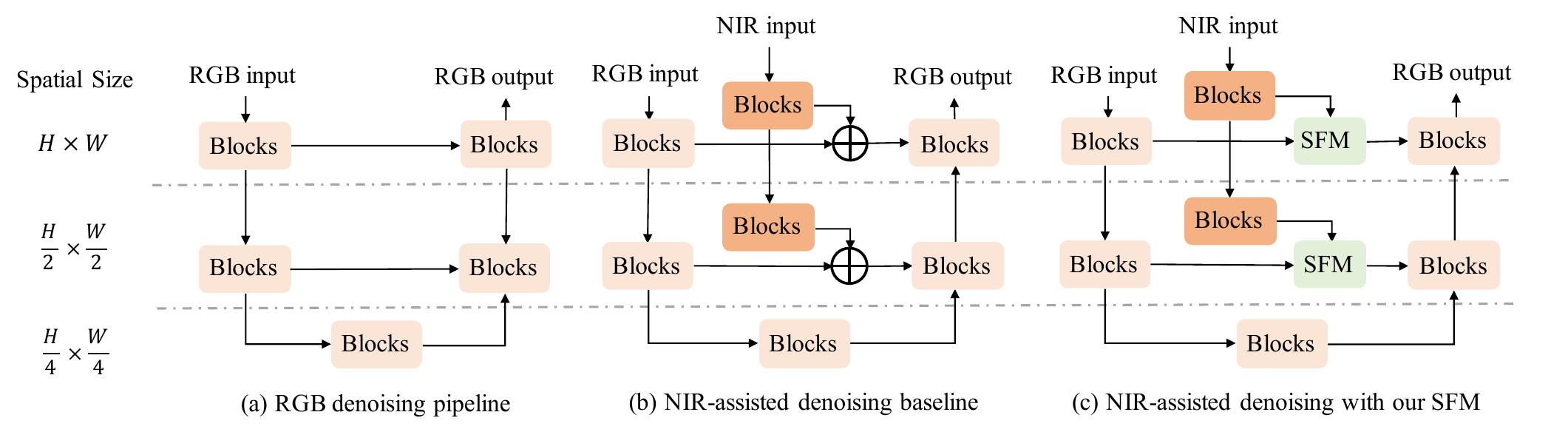}
    \vspace{-4mm}
    \caption{Comparison of different image denoising manners using multi-scale architecture. ($\mathbf{a}$) RGB image denoising. ($\mathbf{b}$) NIR-assisted RGB image denoising baseline. ($\mathbf{c}$) NIR-assisted RGB image denoising with our proposed Selective Fusion Module (SFM). }
    \vspace{-1mm}
    \label{fig:pipeline}
\end{figure*}

NIR-Assisted Image Denoising (NAID) aims at restoring the clean RGB image $\mathbf{\hat{I}} \in \mathbb{R}^{H \times W \times 3}$ from its noisy RGB observation $\mathbf{I_R} \in \mathbb{R}^{H \times W \times 3}$ with the assistance of the NIR image $\mathbf{I_N} \in \mathbb{R}^{H \times W \times 1}$, where $H$ and $W$ denote the height and the width of images, respectively.
Compared to the vanilla image denoising based on the multi-scale encoder-decoder architectures as shown in Fig.~\ref{fig:pipeline} (a), it further utilizes the corresponding NIR image to guide the noise removal.
And this is also the core of the NAID. 
Assuming that the clean NIR images are perfectly consistent with the noisy RGB ones in color and structure, we can simply adapt the existing denoising architectures in  Fig.~\ref{fig:pipeline} (a) to Fig.~\ref{fig:pipeline} (b).
The output $\mathbf{\hat{I}}$ can be written as, 
\begin{equation}\label{naive_fusion}
\mathbf{\hat{I}} = \mathcal{D} ( \mathcal{E}_{N} (\mathbf{I_N}) + \mathcal{E}_{R} (\mathbf{I_R}) ),
\end{equation}
where $\mathcal{D}$ denotes the decoder of the denoising network, $\mathcal{E}_{N}$ and $\mathcal{E}_{R}$ denote the feature encoders for NIR and RGB images, respectively.
However, in practical scenarios, there exist color and structure inconsistencies between the NIR  and the RGB images, as illustrated in Fig.~\ref{fig:structure_example}.
Leveraging the NIR images in a naive way like Eqn.~(\ref{naive_fusion}) only gains limited performance improvement. 

In this work, we suggest alleviating the adverse effects of the inconsistency issue before the combinations of NIR and RGB images.
In particular, we propose a Selective Fusion Module (SFM). SFM predicts and assigns soft weights to NIR and RGB features, thus preparing for subsequent feature fusion. 
Eqn.~\eqref{naive_fusion} can be modified to,
\begin{equation}\label{eqn:sfm}
\mathbf{\hat{I}} = \mathcal{D} ( \mathcal{SFM} (\mathcal{E}_{N} (\mathbf{I_N}) , \mathcal{E}_{R} (\mathbf{I_R}) )).
\end{equation}
Details can be found in the following subsections.

\begin{figure*}[t!]
    \centering
    \includegraphics[width=0.95\linewidth]{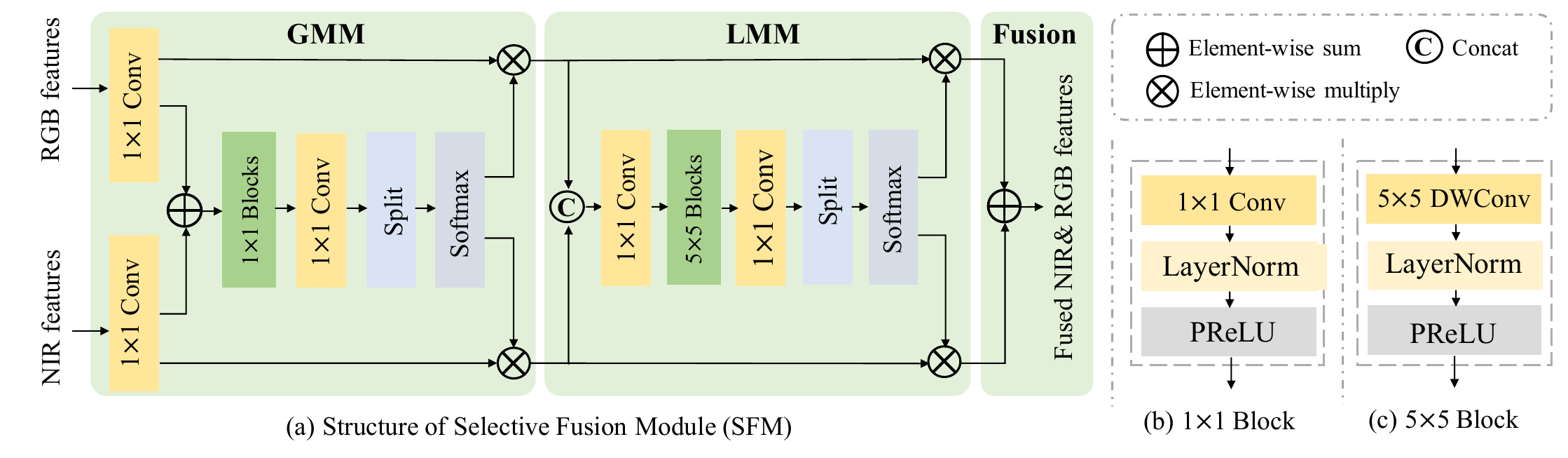}
    \vspace{-4mm}
    \caption{The structure of our proposed Selective Fusion Module (SFM), where Global Modulation Module (GMM) and Local Modulation Module (LMM) focus on color and structure discrepancy issues between the NIR images and RGB ones, respectively. Two $1 \times 1$ blocks and $5 \times 5$  blocks are used in GMM and LMM, respectively.  }
    \label{fig:sfm}
    \vspace{-1mm}
\end{figure*}

\subsection{Selective Fusion Module}
SFM should select valuable information and avoid harmful one from the current NIR-RGB features for feature fusion.
To achieve that, we suggest that SFM predicts and assigns pixel-wise weights for NIR-RGB features fusion. 
Denote the current NIR and RGB features from the corresponding encoders by $\mathbf{F_{N}}$ and $\mathbf{F_{R}}$, SFM can be written as,
\begin{equation}
    \mathcal{SFM} (\mathbf{F_{N}} , \mathbf{F_{R}} ) = \mathbf{W_N} \odot \mathbf{F_{N}} + \mathbf{W_R} \odot  \mathbf{F_{R}},
\end{equation}
where $\odot$ is the pixel-wise multiply operation. $\mathbf{W_N}$ and $\mathbf{W_R}$ denote the weights of NIR and RGB features, respectively.
In order to model the color and structure discrepancy respectively, we decouple the weight $\mathbf{W}$ (including $\mathbf{W_N}$ and $\mathbf{W_R}$) into global and local component, \ie, $\mathbf{W} =  \mathbf{W}^{g}  \odot \mathbf{W}^{l}$, where the former one concentrates on the differences in global information and the latter one focuses on the discrepancy in local information between NIR-RGB features.
Based on that, we further present a Global Modulation Module (GMM) to estimate $\mathbf{W}^{g}$ and a Local Modulation Module (LMM) to estimate $\mathbf{W}^{l}$, as shown in Fig.~\ref{fig:sfm}. 

{\bf Global Modulation Module.} 
\changes{GMM should alleviate the adverse effects of color and luminance inconsistencies on NIR-RGB feature fusion.}
As shown in Fig.~\ref{fig:sfm} (a), it takes the current NIR features $\mathbf{F_{N}}$ and the RGB ones $\mathbf{F_{R}}$ as inputs to estimate the NIR global modulation weights $\mathbf{W}^{g}_\mathbf{N}$ and the RGB ones $\mathbf{W}^{g}_\mathbf{R}$.
Detailly, it first concatenates $\mathbf{F_{N}} \in \mathbb{R}^{C \times H \times W}$ and $\mathbf{F_{R}} \in \mathbb{R}^{C \times H \times W}$ as inputs, generating initial weights $\mathbf{W_{init}^{g}} \in \mathbb{R}^{2C \times H \times W}$ by a $1\times1$ convolutional layer, two $1\times1$ blocks and another $1\times1$ convolutional layer. It can be written as,
$$\mathbf{W_{init}^{g}} = \mathrm{Conv_{1\times1}}(\mathrm{Blocks_{1\times1}}(\mathrm{Conv_{1\times1}}([\mathbf{F_{N}},\mathbf{F_{R}}]))),$$
where $\mathrm{Conv_{1\times1}}$ denotes the $1\times1$ convolutional layer and $\mathrm{Blocks_{1\times1}}$ denotes two $1\times1$ blocks. 
Each $1 \times 1$ block is composed of a $1 \times 1$ convolutional layer, a Layer Normalization~\cite{ba2016layer}, and a PReLU~\cite{he2015delving} function, as shown in Fig.~\ref{fig:sfm} (b).
Next, $\mathbf{W_{init}^{g}}$ is split along channel dimension to get NIR weight maps $\mathbf{\tilde{W}_{N}^{g}} \in \mathbb{R}^{C \times H \times W}$ and RGB ones $\mathbf{\tilde{W}_{R}^{g}} \in \mathbb{R}^{C \times H \times W}$.
Then, a channel-wise softmax operation is applied to $\mathbf{\tilde{W}_{N}^{g}}$ and  $\mathbf{\tilde{W}_{R}^{g}}$, getting the final NIR modulation weights $\mathbf{W}^g_\mathbf{N} \in \mathbb{R}^{C \times H \times W}$ and RGB ones $\mathbf{W}^g_\mathbf{R} \in \mathbb{R}^{C \times H \times W}$, which can be written as,
\begin{equation}
\begin{split}
[\mathbf{W}^g_\mathbf{N}]_{i} = \frac{\exp([\mathbf{\tilde{W}}^g_\mathbf{N}]_{i})}{\exp([\mathbf{\tilde{W}}^g_\mathbf{N}]_{i}) + \exp([\mathbf{\tilde{W}}^g_\mathbf{R}]_{i})},
\\
[\mathbf{W}^g_\mathbf{R}]_{i} = \frac{\exp([\mathbf{\tilde{W}}^g_\mathbf{R}]_{i})}{\exp([\mathbf{\tilde{W}}^g_\mathbf{N}]_{i}) + \exp([\mathbf{\tilde{W}}^g_\mathbf{R}]_{i})}.
\end{split}
\end{equation}
where $[\ \cdot\ ]_{i}$ denotes the selection of the $i$-th channel.
Finally, we modulate the NIR features $\mathbf{F}_\mathbf{N}$ and the RGB ones $\mathbf{F}_\mathbf{R}$ with $\mathbf{W}^g_\mathbf{N}$ and $\mathbf{W}^g_\mathbf{R}$, respectively, \ie, 
\begin{equation}
\mathbf{F}^g_\mathbf{N} = \mathbf{W}^g_\mathbf{N} \odot \mathbf{F_{N}},
\quad
\mathbf{F}^g_\mathbf{R} = \mathbf{W}^g_\mathbf{R} \odot \mathbf{F_{R}}, 
\end{equation}
where $\mathbf{F}^g_\mathbf{N} \in \mathbb{R}^{C \times H \times W}$ and $\mathbf{F}^g_\mathbf{R} \in \mathbb{R}^{C \times H \times W}$ are the globally modulated NIR features and the RGB ones, respectively. 

{\bf Local Modulation Module.}
The Local Modulation Module (LMM) should focus on the structure inconsistency between NIR images and RGB ones. 
\changes{We suggest increasing the receptive field to perceive more local structure information from a range of neighboring pixels, which is better for predicting the fusion weights of NIR and RGB features.}
In detail, LMM takes the globally modulated NIR features $\mathbf{F}^g_\mathbf{N}$ and the RGB ones $\mathbf{F}^g_\mathbf{R}$ as inputs to estimate the local NIR weights $\mathbf{W}^{l}_\mathbf{N}$ and the RGB ones $\mathbf{W}^{l}_\mathbf{R}$,  as shown in Fig.~\ref{fig:sfm} (a).
Without complex network design, LMM is built upon GMM  by replacing the $1 \times 1$ convolutional layer in $1 \times 1$ block to a large kernel depth-wise convolutional layer (DWConv)~\cite{howard2017mobilenets} for capturing more local information, as shown in Fig.~\ref{fig:sfm} (c).
Finally, $\mathbf{W}^l_\mathbf{N}$ and $\mathbf{W}^l_\mathbf{R}$ are employed to get the fused NIR and RGB feature $\mathbf{F_{NR}}$ as,
\begin{equation}
    \mathbf{F_{NR}}  = \mathbf{W}^l_\mathbf{N} \odot  \mathbf{F}^g_\mathbf{N} + \mathbf{W}^l_\mathbf{R} \odot \mathbf{F}^g_\mathbf{R}.
\end{equation}
$\mathbf{F_{NR}}$ is then passed to the decoder to output the final denoising result.

{\bf Discussion.}
As a problem-orientated design module, SFM can make better use of the supplementary information in NIR images, resulting in significantly improved performance on low-light RGB image denoising compared to the naive way (see Fig.~\ref{fig:pipeline} (b)).
In addition, to take full advantage of advanced denoising network architectures~\cite{wang2022uformer,zamir2022restormer,NAFNet,xformer}, we hope SFM can be easily and efficiently integrated into varying denoising backbones.  
Therefore, we intentionally avoid adopting computationally complex operations, and design it as simply and efficiently as possible.
The compact and lightweight architecture makes it only add few parameters and computation costs. 
The related experiment results are presented in Sec.~\ref{sec:experiment}.

\begin{figure*}[t!]
    \centering
     \begin{overpic}[width=0.99\textwidth,grid=False]{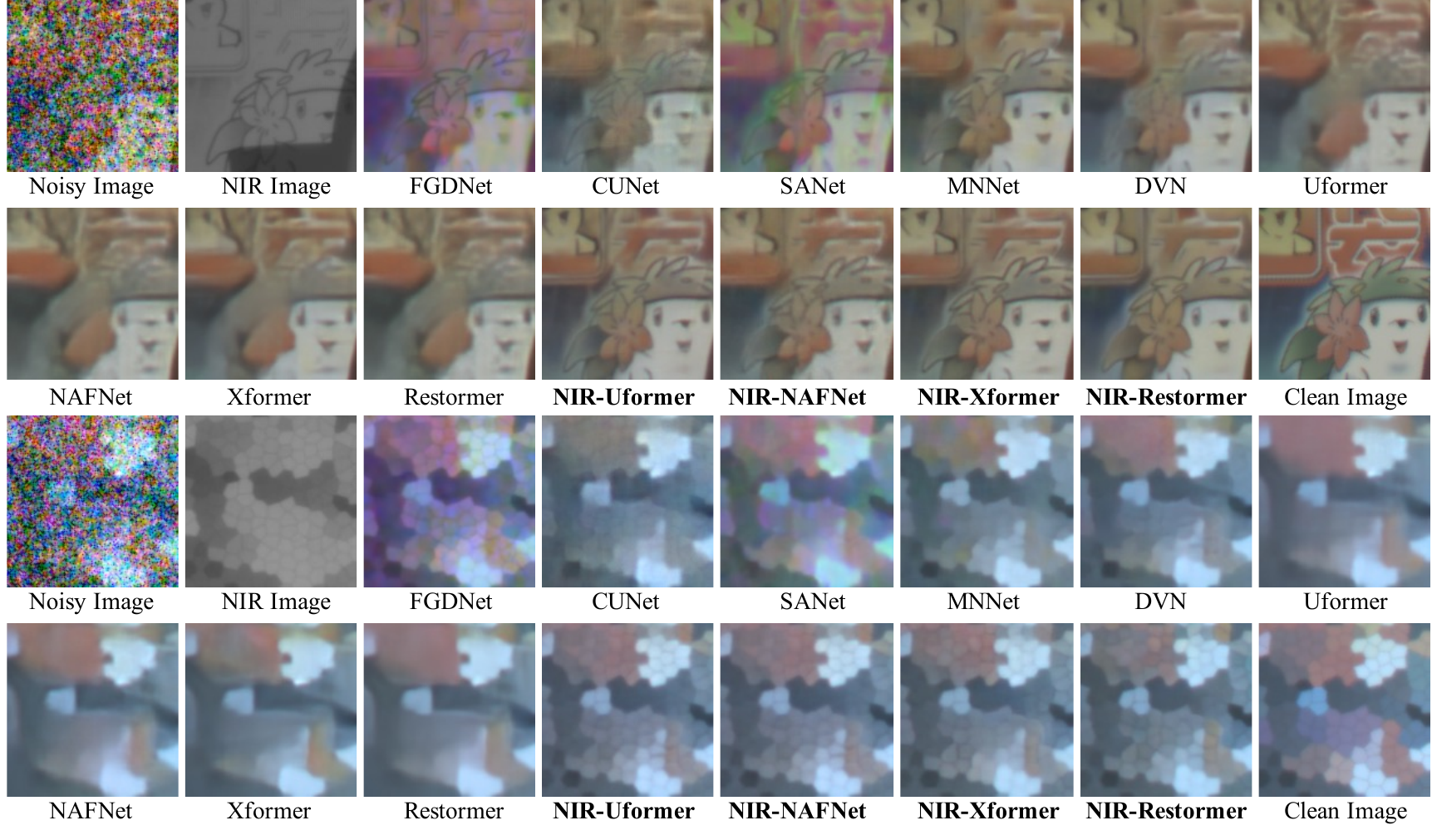}
      \end{overpic}
      \vspace{-4mm}
    \caption{Qualitative comparison on synthetic DVD dataset. ~\textbf{Bold} marks our methods.}
    \label{fig:dvdcomparisons}
\end{figure*}

\begin{table*}[t!] 
    \small
    \renewcommand\arraystretch{1}
    \begin{center}
	\caption{Quantitative comparison on synthetic DVD dataset. \textbf{Bold} marks best results.}
	\label{tab:DVD_table}
        \vspace{-2mm}
        \scalebox{1.05}{
	\begin{tabular}{clccccc}
		\toprule
		\multicolumn{1}{c}{} & 
            \multirow{2}*{Methods} & \multicolumn{1}{c}{$\sigma = 4$} & \multicolumn{1}{c}{$\sigma = 8$} 
            & \multirow{2}*{\tabincell{c}{\#FLOPs \\ (G)}}
            & \multirow{2}*{\tabincell{c}{Time \\ (ms)}}
            & \multirow{2}*{\changes{\tabincell{c}{\#Params \\ (M)}}}\\
		\cmidrule(lr){3-3}\cmidrule(lr){4-4}
		\multicolumn{2}{c}{} & PSNR$\uparrow$/ SSIM$\uparrow$/ LPIPS$\downarrow$ & PSNR$\uparrow$ / SSIM$\uparrow$ / LPIPS$\downarrow$  \\
		\midrule
        \multirow{4}{*}{\begin{tabular}[c]{@{}c@{}} Single-Image\\Denoising \end{tabular}} 
                & Uformer (CVPR'22) & 29.58 / 0.8967 / 0.271 & 27.36 / 0.8632 / 0.352  & 19.16 & 1748 &\changes{20.73}\\
        \multicolumn{1}{c}{} 
                & Restormer (CVPR'22) & 29.67 / 0.9038 / 0.262 & 27.41 / 0.8741 / 0.343  & 70.59 & 2048 & \changes{26.13}\\
        \multicolumn{1}{c}{} 
                & NAFNet (ECCV'22)  & 29.49 / 0.8959 / 0.263 & 27.29 / 0.8638 / 0.336  & 8.10 & 312 & \changes{29.17}\\
        \multicolumn{1}{c}{} 
                & Xformer (ICLR'24)  & 29.57 / 0.8997 / 0.265 & 27.32 / 0.8690 / 0.347  & 73.31 & 5257 & \changes{25.23}\\
            \midrule
		\multirow{9}{*}{\begin{tabular}[c]{@{}c@{}} NIR-Assisted\\ Image \\Denoising \end{tabular}} 
                & FGDNet (TMM'22)  & 23.91 / 0.8371 / 0.439 & 22.02 / 0.7374 / 0.436  & 38.67 & 479 & \changes{1.63}\\
            \multicolumn{1}{c}{} 
		      & SANet (CVPR'23)  & 27.68 / 0.8648 / 0.343 & 25.28 / 0.8304 / 0.413  & 161.06 & 2763 & \changes{5.37}\\
            \multicolumn{1}{c}{} 
		      & CUNet(TPAMI'20)  & 28.01 / 0.8558 / 0.332 & 26.07 / 0.8182 / 0.412  & 14.48 & 542 & \changes{0.44} \\
            \multicolumn{1}{c}{} 
		      & MNNet (IF'22)  & 28.48 / 0.8994 / 0.274 & 26.33 / 0.8697 / 0.353  & 23.68 & 1360 & \changes{0.76}\\
            \multicolumn{1}{c}{} 
		      & DVN (AAAI'22)  & 29.69 / 0.9062 / 0.236 & 27.43 / 0.8799 / 0.292 & 104.05 & 761 & \changes{11.73}\\
		\cmidrule(lr){2-7}
		\multicolumn{1}{c}{} & NIR-Uformer (Ours)  
                & 30.10 / 0.9188 / \textbf{0.192} & 28.03 / \textbf{0.9008} / 0.238 & 24.85 & 2500 & \changes{24.51}\\
		\multicolumn{1}{c}{} & NIR-Restormer (Ours)  
                & 30.22 / \textbf{0.9209} / 0.193 & \textbf{28.11} / 0.8701 / 0.260  & 89.17 & 2747 & \changes{30.26}\\
            \multicolumn{1}{c}{} & NIR-NAFNet (Ours)
                & 30.08 / 0.9005 / 0.208 &    27.86 / 0.8664 / 0.273  & 13.17& 462 & \changes{35.06}\\
            \multicolumn{1}{c}{} & NIR-Xformer (Ours) 
                & \textbf{30.29} / 0.9152 / \textbf{0.192} &    28.09 / 0.8947 / \textbf{0.230}  & 95.63 & 7884 & \changes{30.92}\\
		\bottomrule
	\end{tabular}}
    \end{center}
    \vspace{-1mm}
\end{table*}

\subsection{Loss Function}

We deploy SFM in each scale in multi-scale denoising architecture.
To make SFM play a better role in every scale, we adopt a multi-scale loss function to update network parameters, rather than a naive $\ell_1$ or $\ell_2$ loss. 
It can be written as, 
\begin{equation}\label{multiscale_loss}
\mathcal{L}= \sum_{s=1}^{S-1}|| \mathbf{\hat{I}}_{s} - \mathbf{I}_{\downarrow2^{s-1}} ||_2.
\end{equation}
Therein, S represents the number of network's scales. $\mathbf{I}_{\downarrow2^{s-1}}$ denotes the target image after $\times 2^{s-1}$ down-sampling the ground truth.
The output $\mathbf{\hat{I}}_{s}$ at scale $s$ is generated from a $3 \times 3$ convolutional layer employed after the corresponding decoder, where $\mathbf{\hat{I}}_{1}$ represents the final result with full resolution.

\section{Experiments}\label{sec:experiment}

\subsection{Experimental Settings}\label{5.1}

{\bf Datasets.} 
Experiments are conducted on the synthetic and our Real-NAID datasets.
The details of the Real-NAID dataset can be seen in Sec.~\ref{section:RealNAID_dataset}.  
In addition, We use the DVD~\cite{jin2022darkvisionnet} dataset to generate synthetic noisy images. 
It comprises 307 pairs of clean RGB images (and corresponding RAW images) and NIR images.
267 pairs are used for training and 40 pairs are for testing.
The way to simulate noisy data follows DVD~\cite{jin2022darkvisionnet}.
We first scale the mean value of the clean RAW images, getting synthetic low-light clean RAW images.
Then we add Gaussian noise with the variance $\sigma$ and Poisson noise with a noise level  $\sigma$ to the generated low-light images.
Finally, the synthetic low-light noisy RAW images are converted to RGB ones for training models.
We conduct experiments with $ \sigma  =4$ and $ \sigma = 8$ (the larger  $\sigma$, the heavier noise).

\begin{figure*}[h!]
    \centering
    \includegraphics[width=0.99\linewidth]{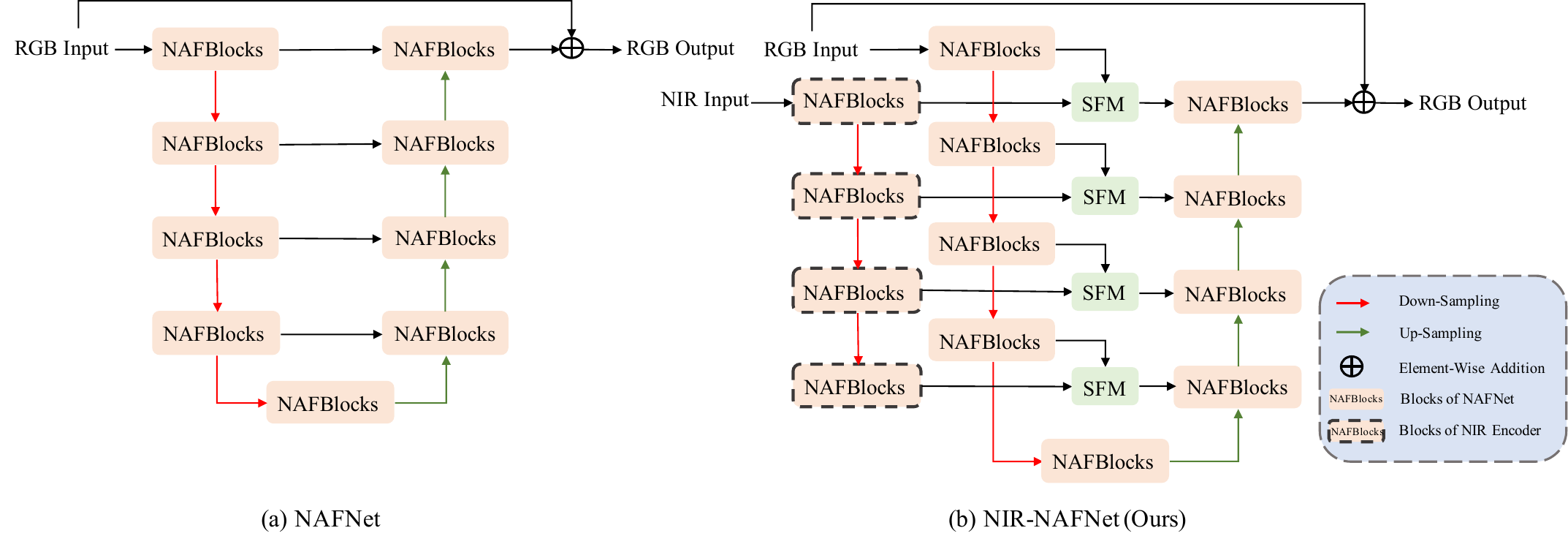}
    \vspace{-3mm}
    \caption{\changes{(a) Single image denoising with NAFNet~\cite{NAFNet}. (b) NIR-assisted image denoising with our NIR-NAFNet.}}
    \label{fig:naf_with_sfm}
\end{figure*}

\begin{figure*}[h!]
    \centering
    \includegraphics[width=0.99\linewidth]{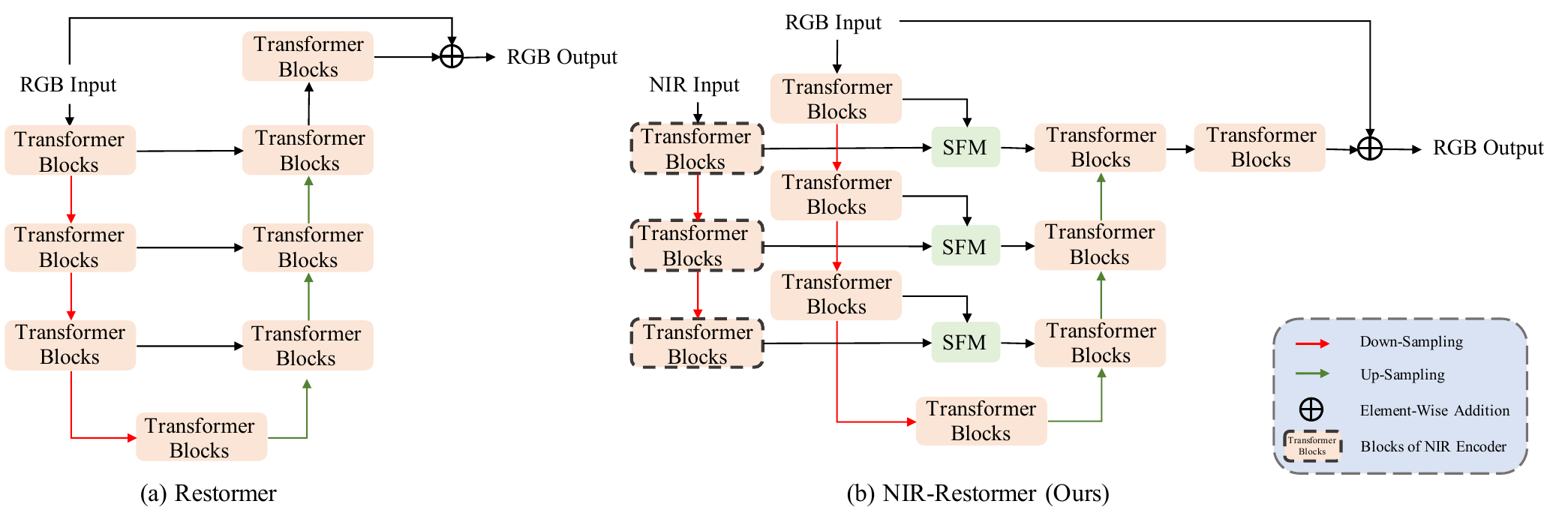}
    \vspace{-3mm}
    \caption{\changes{(a) Single image denoising with Restormer~\cite{zamir2022restormer}. (b) NIR-assisted image denoising with our NIR-Restormer.}}
    \label{fig:restormer_with_sfm}
\end{figure*}

\begin{figure*}[h!]
    \centering
    \includegraphics[width=0.99\linewidth]{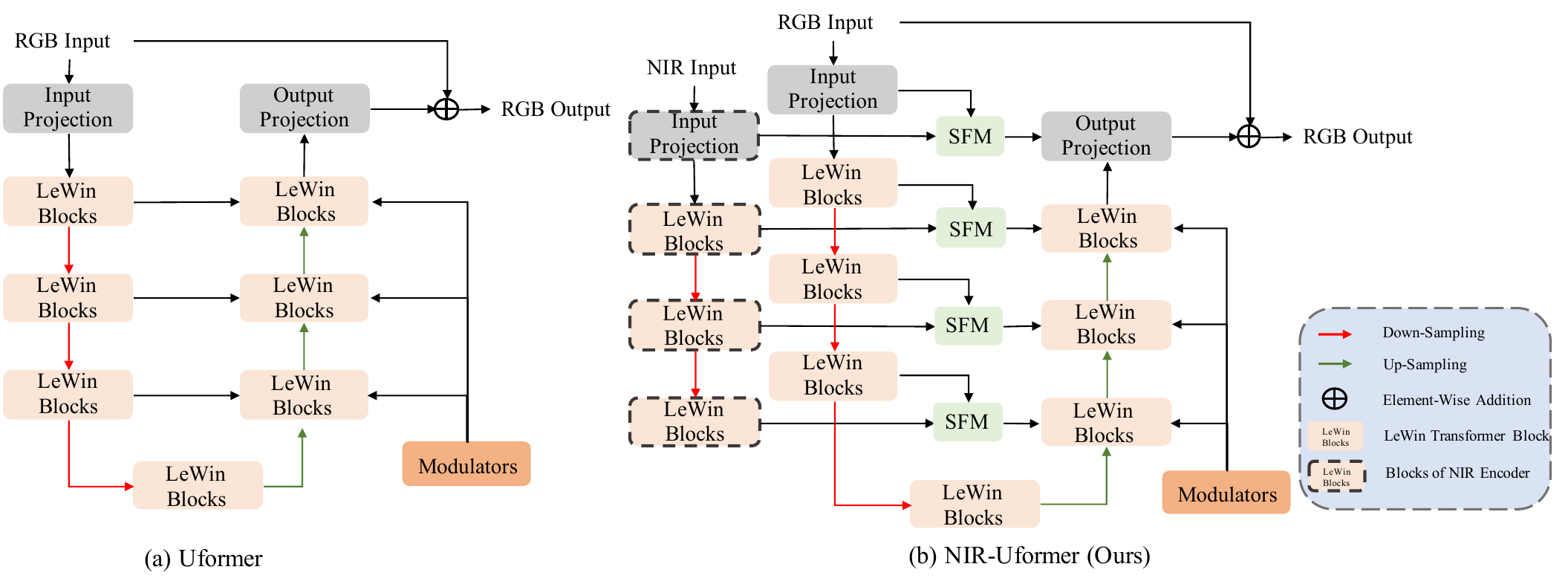}
    \vspace{-3mm}
    \caption{\changes{(a) Single image denoising with Uformer~\cite{wang2022uformer}. (b) NIR-assisted image denoising with our NIR-Uformer.}}
    \label{fig:uformer_with_sfm}
\end{figure*}

\begin{figure*}[h!]
    \centering
    \includegraphics[width=0.99\linewidth]{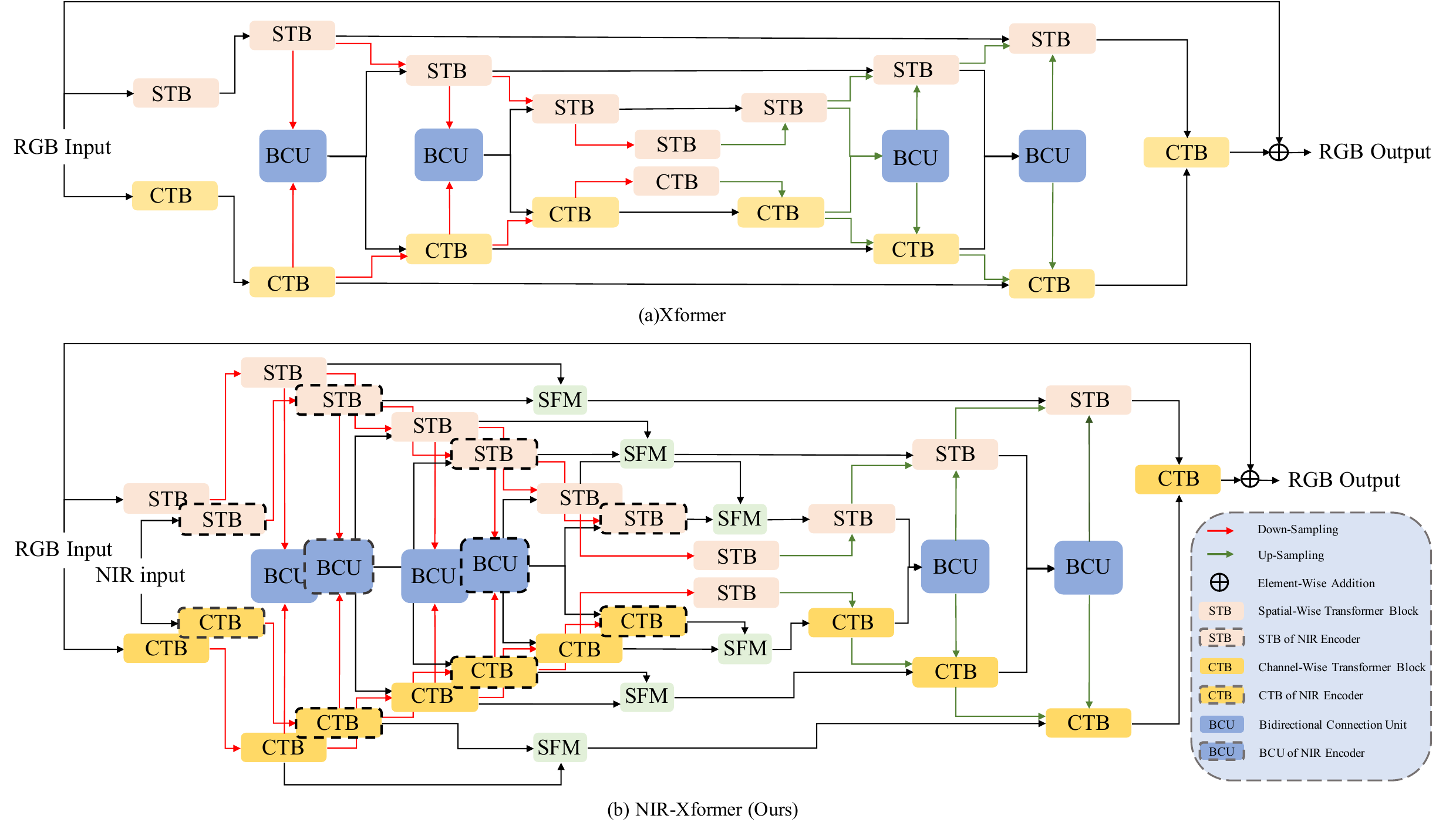}
    \vspace{-3mm}
    \caption{\changes{(a) Single image denoising with Xformer~\cite{xformer}. (b) NIR-assisted image denoising with our NIR-Xformer.}}
    \label{fig:xformer_with_sfm}
\end{figure*}

{\bf Implementation Details.} 
We build our NIR-assisted denoising models by incorporating the proposed SFM into a CNN-based advanced denoising network (\ie, NAFNet~\cite{NAFNet}) and three Transformer-based ones (\ie, Uformer~\cite{wang2022uformer}, Restormer~\cite{zamir2022restormer} and Xformer~\cite{xformer}), which are dubbed \textbf{NIR-NAFNet}, \textbf{NIR-Uformer}, \textbf{NIR-Restormer}, and \textbf{NIR-Xformer}, respectively.  
\changes{Generally, based on the existing single image denoising network, we first add a new encoder branch for processing NIR images, and then utilize the proposed SFM to fuse the same scale NIR and RGB features.
Finally, the fused features are fed into the corresponding decoder for reconstructing clean images.
Network architectures are shown in Fig.~\ref{fig:naf_with_sfm}, Fig.~\ref{fig:restormer_with_sfm}, Fig.~\ref{fig:uformer_with_sfm} and Fig.~\ref{fig:xformer_with_sfm}.
}
All models are trained by the Adam~\cite{Adam} optimizer with $\beta_1$ = 0.9 and $\beta_2$ = 0.999 for 120k iterations.
The batch size is set to 32 and the patch size is set to $128 \times 128$.
For synthetic image denoising, the cosine annealing strategy~\cite{cosine} is employed to steadily decrease the learning rate from $2 \times 10^{-4}$ to $1 \times 10^{-6}$.
For real-world image denoising, the initial learning rate is set to $3\times10^{-4}$ and halved every 20k iterations.
All experiments are conducted with PyTorch~\cite{paszke2019pytorch} on an Nvidia GeForce RTX A6000 GPU.
\vspace{3mm}
\vspace{3mm}
%
\subsection{Comparison with State-of-the-Art Methods}
Experiments are conducted by comparing our NIR-NAFNet, NIR-Uformer, NIR-Restormer, and NIR-Xformer with 9 models, including 4 single image denoising methods (\ie NAFNet~\cite{NAFNet}, Uformer~\cite{wang2022uformer}, Restormer~\cite{zamir2022restormer}, and Xformer~\cite{xformer}) and 5 NIR-assisted image denoising methods (\ie FGDNet~\cite{FGDNet}, SANet~\cite{SANet}, CUNet~\cite{CUNet}, MNNet~\cite{MNNet}, and DVN~\cite{jin2022darkvisionnet}).
To evaluate the performance quantitatively, we calculate three metrics on the RGB channels, 
\ie, Peak Signal to Noise Ratio (PSNR), Structural Similarity (SSIM)~\cite{ssim} and Learned Perceptual Image Patch Similarity (LPIPS)~\cite{lpips}.
The number of parameters (\#Params), \#FLOPs when processing a $128 \times 128$ patch and the inference time when feeding a $1792 \times 1008$ image are reported.

\vspace{10mm}
\begin{table*}[t!] 
    \small
    \setlength{\tabcolsep}{6pt}
    \renewcommand\arraystretch{1}
    \begin{center}
	\caption{Quantitative comparison on our Real-NAID dataset. \textbf{Bold} marks best results.}
	\label{tab:RealNAID_table}
        \vspace{-2mm}
        \scalebox{1}{
	\begin{tabular}{clcccc}
		\toprule
            \multicolumn{1}{c}{}
            & \multirow{2}*{Methods} & \multicolumn{1}{c}{Low-Level Noise} 
            & \multicolumn{1}{c}{Middle-Level Noise} 
            & \multicolumn{1}{c}{High-Level Noise}\\
		\cmidrule(lr){3-3}\cmidrule(lr){4-4}\cmidrule(lr){5-5}
		\multicolumn{2}{c}{} & PSNR$\uparrow$ / SSIM$\uparrow$ / LPIPS$\downarrow$ 
            & PSNR$\uparrow$ / SSIM$\uparrow$ / LPIPS$\downarrow$
            & PSNR$\uparrow$ / SSIM$\uparrow$ / LPIPS$\downarrow$ \\
		\midrule
		\multirow{4}{*}{\begin{tabular}[c]{@{}c@{}} Single-Image\\Denoising \end{tabular}} 
                & Uformer (CVPR'22) & 25.56 / 0.7736 / 0.304 
                & 24.52 / 0.7418 / 0.347 & 23.31 / 0.7091 / 0.389 \\
		\multicolumn{1}{c}{} 
                & Restormer (CVPR'22) & 25.89 / 0.7842 / 0.294 
                & 24.98 / 0.7572 / 0.333 & 23.82 / 0.7297 / 0.387 \\
		\multicolumn{1}{c}{} 
                & NAFNet (ECCV'22)  & 25.71 / 0.7780 / 0.294 
                & 24.76 / 0.7482 / 0.335 & 23.71 / 0.7186 / 0.378\\
		\multicolumn{1}{c}{} 
                & Xformer (ICLR'24)  &  25.91 / 0.7804 / 0.292
                & 24.81 / 0.7496 / 0.338 & 23.73 / 0.7201 / 0.382\\
            \midrule
		\multirow{9}{*}{\begin{tabular}[c]{@{}c@{}} NIR-Assisted\\Image \\ Denoising \end{tabular}} 
                & FGDNet (TMM'22)  & 24.25 / 0.7676 / 0.368 
                & 22.89 / 0.7367 / 0.430 & 21.86 / 0.7080 / 0.509 \\
            \multicolumn{1}{c}{} 
		      & CUNet (TPAMI'20)  & 24.05 / 0.7314 / 0.313 
                & 23.29 / 0.7031 / 0.380 & 22.41 / 0.6398 / 0.449 \\
            \multicolumn{1}{c}{} 
		      & SANet (CVPR'23) & 24.93 / 0.7679 / 0.359 
                & 23.74 / 0.7335 / 0.416 & 22.69 / 0.7028 / 0.476 \\
            \multicolumn{1}{c}{} 
		      & MNNet (IF'22)  & 25.68 / 0.7797 / 0.313 
                & 24.64 / 0.7512 / 0.364 & 23.36 / 0.7194 / 0.419 \\
            \multicolumn{1}{c}{} 
		      & DVN (AAAI'22) & 25.96 / 0.7853 / 0.298 
                & 24.93 / 0.7578 / 0.332 & 23.95 / 0.7360 / 0.382 \\
            \cmidrule(lr){2-5}
                & NIR-Uformer (Ours) & 25.91 / 0.7919 / 0.276 
                & 25.14 / 0.7714 / 0.299 & 24.28 / 0.7534 / 0.321 \\
		\multicolumn{1}{c}{} 
                & NIR-Restormer (Ours)  & \textbf{26.22} / \textbf{0.7963} / 0.265 
                & \textbf{25.51 / 0.7767 / \textbf{0.293}} & \textbf{24.76 / 0.7626 / 0.315} \\
    	\multicolumn{1}{c}{} 
                & NIR-NAFNet (Ours) & 26.06 / 0.7905 / 0.274
                & 25.26 / 0.7676 / 0.303 & 24.48 / 0.7503 / 0.321 \\
    	\multicolumn{1}{c}{} 
                & NIR-Xformer (Ours) & 26.19 / 0.7959 / \textbf{0.264} 
                & 25.38 / 0.7729 / \textbf{0.293} & 24.66 / 0.7595 / 0.319 \\
		\bottomrule
	\end{tabular}}
    \end{center}
    \vspace{-1mm}
\end{table*}

{\bf Results on Synthetic DVD dataset.}
The quantitative results on the synthetic DVD dataset are shown in Table~\ref{tab:DVD_table}.
The best results are shown in bold.
It can be observed that our method significantly improves performance against single-image denoising methods, thereby demonstrating the effectiveness of NIR images.
In comparison with existing NIR-assisted denoising ones, our methods also outperform by a large margin, as the proposed SFM  overcomes the discrepancy issues between the NIR-RGB images while coupling with the advanced denoising backbone successfully. 
In particular, our NIR-NAFNet makes a better trade-off between performance and efficiency than other methods.
Besides, The qualitative results in Fig.~\ref{fig:dvdcomparisons} show that our methods restore more realistic textures and fewer artifacts than others.

\begin{figure*}[t!]
    \centering
     \begin{overpic}[width=0.99\textwidth,grid=False]{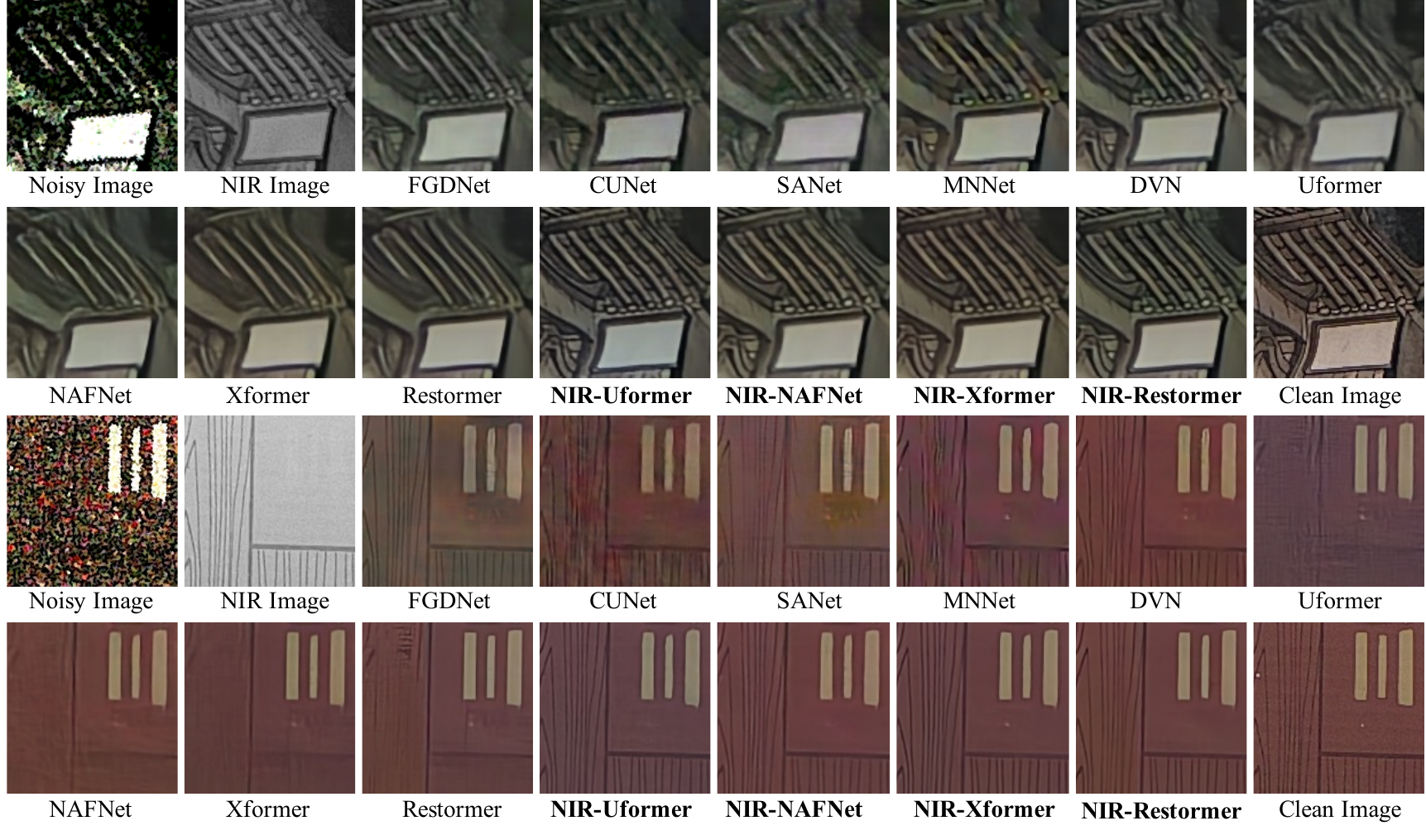}
      \end{overpic}
      \vspace{-4mm}
    \caption{Qualitative comparison on our Real-NAID dataset. ~\textbf{Bold} marks our methods.}
    \label{fig:realnaidcomparisons}
\end{figure*}

{\bf Results on Real-NAID Dataset.}
Real-world data has much more complex degradation than synthetic ones.
The quantitative results in Table~\ref{tab:RealNAID_table} show that our methods still keep high performance in the real world.
Taking NIR-Restormer as an example, our proposed NIR-Restormer achieves 0.33dB, 0.54dB, and 0.94dB PSNR gains than Restormer~\cite{zamir2022restormer} in dealing with low-level, middle-level and high-level noise, respectively.
The higher the level of noise, the greater the improvement achieved by our method, which further indicates the advantage of the utilization of NIR information for low-light noise removal.
The qualitative results in Fig.~\ref{fig:realnaidcomparisons} demonstrate that our models still recover fine-scale details in the real world, while other NIR-assisted denoising methods may produce artifacts.

\section{Ablation Study}\label{sec:ablation}
We conduct ablation studies on our Real-NAID dataset using NIR-NAFNet.
Unless otherwise stated, ablation experiments are evaluated at three noise levels, and the metrics are reported by averaging them on three noise levels.
The experiments include comparisons with NAID baselines, comparisons with more fusion manners, effect of SFM, and effect of multi-scale loss.

\subsection{Comparisons with NAID Baselines }\label{sec:comparison_NAID}
To demonstrate the effectiveness of our  SFM, we compare our methods with NAID baselines using varying backbones.
Specifically, we simply sum the NIR and RGB features as our baseline, as shown in  Fig.~\ref{fig:pipeline} (b).
We utilize three Transformer-based denoising networks (\ie, Uformer~\cite{wang2022uformer}, Restormer~\cite{zamir2022restormer}, Xformer~\cite{xformer}) and a CNN-based one (\ie, NAFNet~\cite{NAFNet}) to conduct experiments, named `Uformer-Baseline',  `Restormer-Baseline', `Xformer-Baseline', and `NAFNet-Baseline', respectively.
\changes{The results are summarized in Table~\ref{tab:baseline_table2}.
First, the proposed SFM improves the performance of all NAID baselines with few computational cost additions.
Second, the performance of the denoising model decreases as the noise level increases.
This is because higher levels of noise cause more deterioration in RGB images, making image restoration more challenging.
Third, the performance improvement from SFM becomes increasingly significant with higher noise levels.
This is because the inconsistency between RGB and NIR images becomes more severe as the RGB image degradation becomes more severe.
The simple summation in NAID baselines ignores it, whereas SFM better allocates fusion weights for RGB and NIR features.
}

\begin{table*}[t!] 
    \renewcommand\arraystretch{1}
    \setlength{\tabcolsep}{4pt}
    \begin{center}
	\caption{Comparisons of our proposed method with the NAID baselines (see Fig.~\ref{fig:pipeline} (b)) on different backbones on our Real-NVID dataset.}
	\label{tab:baseline_table2}
        \vspace{-2mm}
        \scalebox{0.98}{
	\begin{tabular}{cccccccc}
		\toprule
		\multirow{2}*{} & \multirow{2}*{Methods} & \multicolumn{1}{c}{Low-Level Noise} & \multicolumn{1}{c}{Middle-Level Noise} & \multicolumn{1}{c}{High-Level Noise}
            & \multirow{2}*{\changes{\tabincell{c}{\#FLOPs \\ (G)}}}
            & \multirow{2}*{\changes{\tabincell{c}{Time \\ (ms)}}}
            & \multirow{2}*{\changes{\tabincell{c}{\#Params \\ (M)}}}\\
		\cmidrule(lr){3-3}\cmidrule(lr){4-4}\cmidrule(lr){5-5}
		\multicolumn{2}{c}{} & PSNR$\uparrow$/ SSIM$\uparrow$/ LPIPS$\downarrow$ & PSNR$\uparrow$/ SSIM$\uparrow$/ LPIPS$\downarrow$ & PSNR$\uparrow$/ SSIM$\uparrow$/ LPIPS$\downarrow$\\
            \midrule
            \multirow{2}{*}{\begin{tabular}[c]{@{}c@{}} Spatial-Wised\\Transformer-Based \end{tabular}} 
		&  Uformer-Baseline  
                & 25.80 / 0.7904 / 0.269 & 25.03 / 0.7687 / 0.294 
                & 24.02 / 0.7488 / 0.320 
                & \changes{23.61} & \changes{2284} 
                & \changes{23.39}\\
            \multicolumn{1}{c}{} & NIR-Uformer 
                & 25.91 / 0.7917 / 0.276 & 25.14 / 0.7714 / 0.299
                & 24.28 / 0.7534 / 0.321
                & \changes{24.85} & \changes{2500} 
                & \changes{24.51}\\
		\midrule
		\multirow{2}{*}{\begin{tabular}[c]{@{}c@{}} Channel-Wised\\Transformer-Based \end{tabular}} 
            & Restormer-Baseline  
                & 26.12 / 0.7948 / 0.265 & 25.38 / 0.7733 / 0.293
                & 24.61 / 0.7587 / 0.316
                & \changes{86.75} & \changes{2578} 
                & \changes{29.76}\\
		\multicolumn{1}{c}{} & NIR-Restormer  
                & 26.22 / 0.7963 / 0.265 & 25.51 / 0.7767 / 0.293
                & 24.76 / 0.7626 / 0.315
                & \changes{89.17} & \changes{2747} 
                & \changes{30.26}\\
            \midrule
		\multirow{2}{*}{\begin{tabular}[c]{@{}c@{}} Mixed\\Transformer-Based \end{tabular}} 
            & Xformer-Baseline  
                & 26.10 / 0.7927 / 0.264 & 25.23 / 0.7694 / 0.291
                & 24.41 / 0.7553 / 0.314
                & \changes{90.78} & \changes{7616} 
                & \changes{29.91}\\
		\multicolumn{1}{c}{} & NIR-Xformer  
                & 26.19 / 0.7959 / 0.264 & 25.38 / 0.7729 / 0.293
                & 24.66 / 0.7595 / 0.319
                & \changes{95.63} & \changes{7884} 
                & \changes{30.92}\\
            \midrule
            \multirow{2}{*}{CNN-Based} 
		&  NAFNet-Baseline  
                & 25.92 / 0.7883 / 0.277 & 25.00 / 0.7628 / 0.304
                & 24.16 / 0.7429 / 0.329
                & \changes{11.72} & \changes{462} 
                & \changes{34.16}\\
            \multicolumn{1}{c}{} & NIR-NAFNet
                & 26.06 / 0.7905 / 0.274 & 25.26 / 0.7676 / 0.303
                & 24.48 / 0.7503 / 0.321
                & \changes{13.17} & \changes{462} 
                & \changes{35.06}\\
		\bottomrule
	\end{tabular}}
    \end{center}
  \vspace{-1mm}
\end{table*}

\begin{table}[ht] 
    \renewcommand\arraystretch{1.25}
    \setlength{\tabcolsep}{2pt}
    \begin{center}
        \caption{Comparisons with more fusion manners.}
	\label{tab:sfm_table}
        \vspace{-2mm}
        \scalebox{1}{
	\begin{tabular}{c c c c c c}
		\toprule
            Input Images & Fusion  Methods & PSNR$\uparrow$/ SSIM$\uparrow$/ LPIPS$\downarrow$  & Time (ms)\\
            \midrule
            RGB & None & 24.72 / 0.7486 / 0.336 & 312  \\
            RGB and NIR & Sum       & 25.03 / 0.7647 / 0.304   & 461     \\ 
            RGB and NIR & Channel Att.~\cite{channel-attention}     & 25.12 / 0.7664 / 0.309  & 461   \\ 
            \changes{RGB and NIR} &   \changes{MDTA~\cite{zamir2022restormer}}  &  \changes{25.16 / 0.7670 / 0.296}   & \changes{1614}  \\ 
            RGB and NIR & Cross Att.~\cite{swin} & 25.18 / 0.7692 / 0.304   &   4107   \\ 
            RGB and NIR & Ours               & 25.26 / 0.7695 / 0.299   &  462   \\ 
		\bottomrule
	\end{tabular}}
    \end{center}
  \vspace{-5mm}
\end{table}

\subsection{Comparisons with More Fusion Manners}\label{sec:comaprison_fusion}
We compare our selective fusion manner with other fusion methods, including feature summation, channel attention~\cite{channel-attention}, and cross attention~\cite{swin} to demonstrate our method's superiority. 
The results on Real-NAID dataset are shown in Table~\ref{tab:sfm_table}. 
It can be observed the usage of NIR images and other prevailing fusion methods can improve performance to some degree.
However, these methods do not achieve competitive performance compared with our SFM while sometimes introducing higher computation costs.
\changes{
This may be due to methods like MDTA~\cite{zamir2022restormer} or cross attention~\cite{swin} only focus on correlation on the channel dimension or spatial dimension, while our proposed method focuses on both dimensions.}
These illustrate the superiority of the proposed method.

\subsection{Effect of SFM}\label{sec:more_SFM}
To illustrate the effect of SFM, we conduct a more detailed investigation and exploration on GMM and LMM.
\begin{figure*}[t!]
    \centering
     \begin{overpic}[width=0.98\textwidth,grid=False]
     {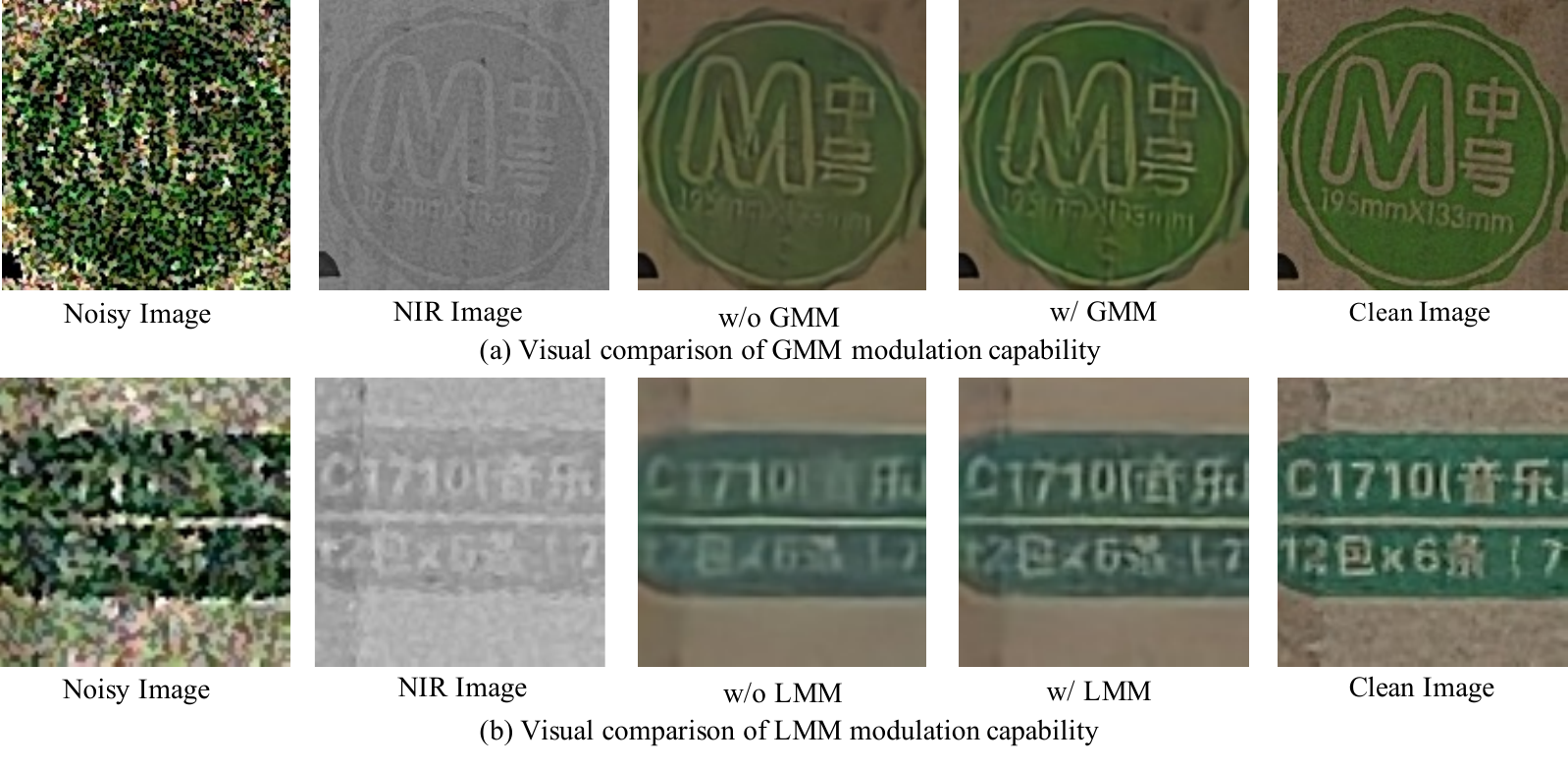}
    \end{overpic}
    \vspace{-4mm}
    \caption{(a) Visual comparison of GMM modulation capability. GMM tends to modulate global color. (b) Visual comparison of LMM modulation capability. LMM tends to modulate local structure.}
    \label{fig:GMM_LMM_visualize}
    \vspace{-2mm}
\end{figure*}
\begin{figure*}[th!]
    \centering
    \includegraphics[width=0.98\linewidth]{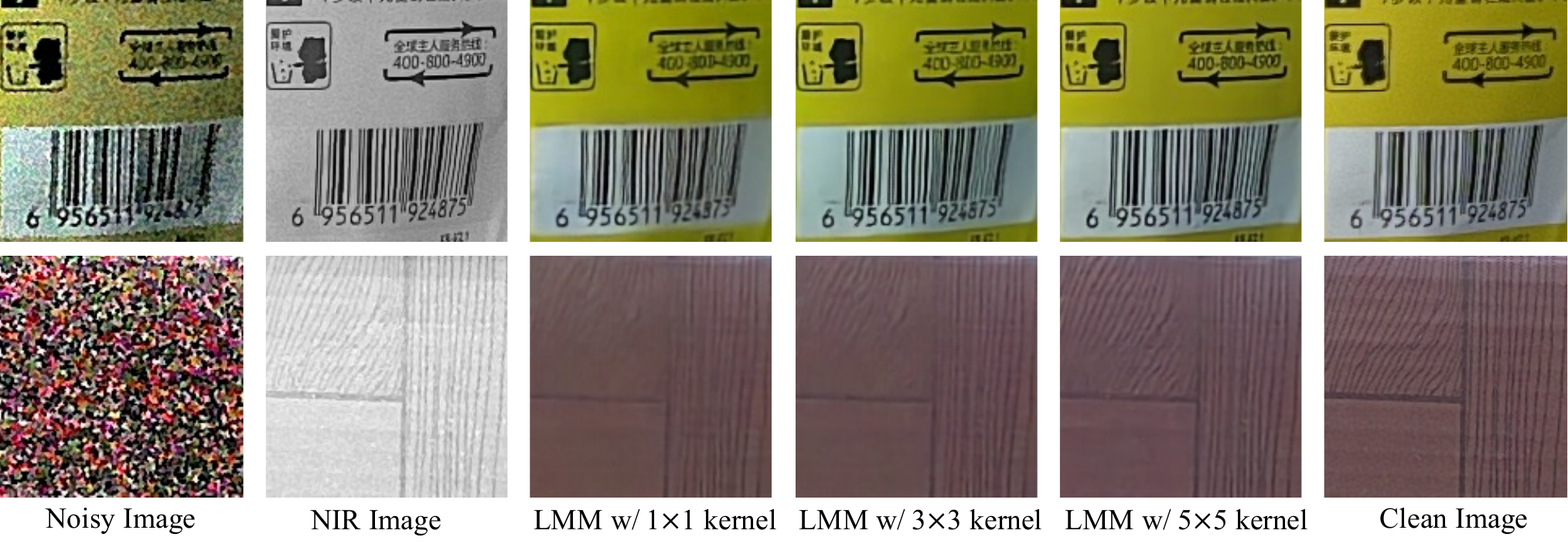}
    \vspace{-2mm}
    \caption{\changes{Visual comparison of different kernel sizes of the DWConvs in our LMM.}}
    \label{fig:ks}
    \vspace{-3mm}
\end{figure*}
{\bf Effect of GMM.}
As shown in Table~\ref{tab:compoents_table}, the incorporation of GMM yields 0.13dB PSNR improvement on its own and yields 0.1dB PSNR improvement on the basis of LMM, which can be attributed to its effective handling of the inconsistency between NIR-RGB images in color.
Additionally, we provide a visual comparison to demonstrate the impact of GMM on visual results.
As shown in Fig.~\ref{fig:GMM_LMM_visualize} (a), the incorporation of our GMM helps better color recovery.

\begin{table}[t!]
    \centering
    \caption{Quantitative comparison with different modulation modules in SFM.}
    \vspace{-2mm}
    \label{tab:compoents_table}
    \setlength{\tabcolsep}{18pt}
    \renewcommand\arraystretch{1.25}
    \scalebox{1}{
        \begin{tabular}{cccc}
            \toprule
                GMM & LMM 
                & PSNR$\uparrow$ / SSIM$\uparrow$ / LPIPS$\downarrow$ \\
            \midrule
            $\times$ & $\times$ & 25.03 / 0.7647 / 0.304\\
            $\checkmark$ & $\times$ & 25.16 / 0.7675 / 0.302 \\
            $\times$ & $\checkmark$ & 25.16 / 0.7653 / 0.302  \\
            $\checkmark$ & $\checkmark$ & 25.26 / 0.7695 / 0.299 \\
            \bottomrule
        \end{tabular}}
\end{table}

\begin{table}[t!] 
    \renewcommand\arraystretch{1.25}
    \setlength{\tabcolsep}{20pt}
    \centering
    \caption{Quantitative comparison of different kernel sizes of DWConv in LMM.}
    \vspace{-2mm}
    \label{tab:ksize_table}
        \scalebox{1}{
	\begin{tabular}{ccccc}
		\toprule
            \tabincell{c}{Kernel Size}  & PSNR$\uparrow$ / SSIM$\uparrow$ / LPIPS$\downarrow$ \\
        \midrule
		$3 \times 3$ & 25.22 / 0.7685 / 0.298 \\
		$5 \times 5$ & 25.26 / 0.7695 / 0.299 \\
		$7 \times 7$ & 25.28 / 0.7717 / 0.301 \\
		\bottomrule
	\end{tabular}}
    \vspace{0mm}
\end{table}

\begin{table}[t!]
    \centering
    \caption{Quantitative comparison of different arrangements of GMM and LMM.}
    \label{tab:composition_table}
    \vspace{-2mm}
    \setlength{\tabcolsep}{20pt}
    \renewcommand\arraystretch{1.25}
    \scalebox{1}{
        \begin{tabular}{ccc}
            \toprule
                \multicolumn{2}{c}{Arrangement} 
                &  PSNR$\uparrow$ / SSIM$\uparrow$ / LPIPS$\downarrow$ \\
            \midrule
            \multicolumn{2}{c}{GMM + GMM} & 25.17 / 0.7652 / 0.302 \\
            \multicolumn{2}{c}{LMM + LMM} & 25.18 / 0.7650 / 0.301 \\
            \multicolumn{2}{c}{LMM + GMM} & 25.19 / 0.7669 / 0.301 \\
            \multicolumn{2}{c}{GMM + LMM} & 25.26 / 0.7695 / 0.299 \\
            \bottomrule
        \end{tabular}}
        \vspace{-1mm}
\end{table}
{\bf Effect of LMM.}
As shown in Table~\ref{tab:compoents_table}, there is a 0.1dB drop on PSNR when removing LMM from SFM. 
We provide a qualitative result to demonstrate its impact on visual results.
As shown in Fig.~\ref{fig:GMM_LMM_visualize} (b), the incorporation of our LMM in SFM helps fine-scale texture recovery.
Moreover, we explore the appropriate receptive field size in LMM by employing varying kernel sizes of DWConv as shown in Table~\ref{tab:ksize_table}.
\changes{Larger convolution kernels have larger receptive fields, thus it is beneficial for local feature extraction~\cite{ks}.
Then, better local feature extraction is conducive to better prediction of the fusion weights of NIR and RGB features, thus making better use of NIR image information.
Finally, in terms of visual effects, it helps better structure recovery, as shown in Fig.~\ref{fig:ks}.}
But it is improved marginally when the kernel size is larger than $5 \times 5$.
For the sake of simplicity and efficiency, we set the kernel size of DWConv to $5 \times 5$ as default. 
\begin{table}[h!]
    \centering
    \caption{\changes{Quantitative comparison of two GMMs or LMMs in one SFM.}}
    \label{tab:more_GMMs_or_more_LMMs}
    \vspace{-2mm}
    \setlength{\tabcolsep}{5pt}
    \renewcommand\arraystretch{1.25}
    \scalebox{1}{
        \begin{tabular}{cccc}
            \toprule
               $\#$GMM in SFM 
                & $\#$LMM in SFM &  PSNR$\uparrow$ / SSIM$\uparrow$ / LPIPS$\downarrow$ \\
            \midrule
            1 & 1 & 25.26 / 0.7695 / 0.299 \\
            2 & 1 & 25.26 / 0.7691 / 0.301 \\
            1 & 2 & 25.26 / 0.7715 / 0.300 \\
            2 & 2 & 25.27 / 0.7711 / 0.299  \\
            \bottomrule
        \end{tabular}}
        \vspace{-1mm}
\end{table}
{\bf Effect of GMM and LMM Arrangement.}
Here, we conduct experiments with different arrangements of GMM and LMM.
The results are shown in Table~\ref{tab:composition_table}.
`GMM + GMM' and `LMM + LMM' mean we modulate features with 2 GMMs and 2 LMMs respectively. They result in limited performance gain.
It shows that our performance improvement is not due to a simple increase in parameter numbers.
`LMM + GMM' means we modulate features first locally and then globally. It also leads to limited improvement.
It may be because the significant difference in global content leads to inaccurate local feature modulation.
Therefore, we deploy a GMM to handle color discrepancy first followed by an LMM dealing with structure discrepancy, dubbed `GMM + LMM', achieving better results.
\changes{Additionally, we conduct experiments on different numbers of GMM and LMM in one SFM.
Table~\ref{tab:more_GMMs_or_more_LMMs} shows more GMMs or LMMs in one SFM only get minor improvement.
For simplicity, We set one GMM and one LMM in one SFM.}

{\bf Effect of Numbers of Blocks in GMM and LMM.}
\changes{We conduct experiments on the effect of the number of blocks in Table~\ref{tab:blocks_table}.
Table~\ref{tab:blocks_table} shows that more blocks in GMM and LMM can both improve performance. 
However, increasing the number of $1 \times 1$ and $5 \times 5$ blocks from one to five only brings a 0.04 dB PSNR improvement, which is minor.
%
%
For the sake of simplicity and efficiency, we only adopt one  $1 \times 1$ block in GMM and one $5 \times 5$ block in LMM.}
{\bf Effect of Number of SFM.}
Here we investigate the effect of incorporating multiple SFMs at each scale of NIR-NAFNet.
The results are shown in Table~\ref{tab:number_table}.
It can be observed that the performance generally increases marginally as the number of SFMs grows.
Also for the sake of simplicity and efficiency, we only set the number of SFM to 1 at each scale.

{\bf Efficiency of GMM and LMM.}
Both GMM and LMM are lightweight modules that do not increase the number of parameters and inference time too much.
The number of parameters of GMM and LMM only account for $1.5\%$ and $1.1\%$ of those of NIR-NAFNet, respectively. 
Applying an SFM on NIR-NAFNet only results in a time increase of 1 ms.

\begin{table}[t!]
    \centering
    \caption{\changes{Quantitative comparison of different numbers of blocks in GMM and LMM.}}
    \label{tab:blocks_table}
    \vspace{-2mm}
    \setlength{\tabcolsep}{5pt}
    \renewcommand\arraystretch{1.25}
    \scalebox{1}{
        \begin{tabular}{ccc}
            \toprule
                $\#$Blocks in GMM 
                & $\#$Blocks in LMM 
                &  PSNR$\uparrow$ / SSIM$\uparrow$ / LPIPS$\downarrow$ \\
            \midrule
            1 & 1 & 25.26 / 0.7695 / 0.299 \\
            3 & 1 & 25.27 / 0.7702 / 0.302 \\
            5 & 1 & 25.27 / 0.7701 / 0.302 \\
            1 & 3 & 25.27 / 0.7709 / 0.299 \\
            1 & 5 & 25.28 / 0.7701 / 0.302 \\
            3 & 3 & 25.28 / 0.7705 / 0.300 \\
            5 & 5 & 25.30 / 0.7718 / 0.302 \\
            \bottomrule
        \end{tabular}}
        \vspace{-1mm}
\end{table}

\begin{table}[t!] 
    \renewcommand\arraystretch{1.25}
    \setlength{\tabcolsep}{20pt}
    \centering
    \caption{Quantitative comparison of different numbers of SFM.}
    \label{tab:number_table}
    \vspace{-2mm}
        \scalebox{1.2}{
    \begin{tabular}{ccc}
        \toprule
        \tabincell{c}{$\#$SFM} & PSNR$\uparrow$ / SSIM$\uparrow$ / LPIPS$\downarrow$ \\
            \midrule
        1 & 25.26 / 0.7695 / 0.299 \\
        3 & 25.28 / 0.7699 / 0.299  \\
        5 & 25.29 / 0.7669 / 0.301  \\
        \bottomrule
    \end{tabular}}
    \vspace{0mm}
\end{table}

\subsection{Effect of Multi-Scale Loss}\label{sec:fusion_methods_comparison}
The multi-scale loss calculates the difference between output and supervision at different scales, and it can slightly improve performance with few increasing training costs.
\changes{Note that we conduct the experiment several times to eliminate randomness. 
The results are consistent, only with a small variance.}
Taking NAFNet~\cite{NAFNet} as an example, we evaluate its effect by comparing it with naive $\ell_2$ loss. Table~\ref{tab:loss_table} shows that it improves the performance of both single-image and NIR-assisted image denoising. 
And the improvement is more obvious for the latter. 
This shows that multi-scale loss can help SFM work better at each scale.
\begin{table}[t!] 
    \setlength{\tabcolsep}{8pt}
    \renewcommand\arraystretch{1.25}
    \begin{center}
        \caption{Comparisons of different loss functions.}
	\label{tab:loss_table}
        \vspace{-2mm}
        \scalebox{1.05}{
	\begin{tabular}{c c c}
		\toprule
            Input Images & Loss Function  & PSNR$\uparrow$/ SSIM$\uparrow$/ LPIPS$\downarrow$ \\ 
            \midrule
            RGB & Naive $\ell_2$ & 24.72 / 0.7476 / 0.343 \\     
            RGB & Multi-Scale & 24.73 / 0.7483 / 0.336      \\
            RGB and NIR & Naive $\ell_2$  & 25.20 / 0.7691 / 0.299 \\ 
            RGB and NIR & Multi-Scale & 25.26 / 0.7695 / 0.299 \\ 
		\bottomrule
	\end{tabular}}
    \end{center}
  \vspace{0mm}
\end{table}

\section{Limitations}
\changes{Denoising models in this work are only trained on images captured by one camera.
Given that the noise models of different cameras may be different, our denoising models may not generalize well on other cameras or devices.
In the future, we consider constructing a larger real-world NIR-RGB dataset covering multiple cameras.
Thus, denoising models trained on this dataset may generalize well.}

\section{Conclusion}
Near-infrared (NIR) images can help restore fine-scale details while removing noise from noisy RGB images, especially in low-light environments. 
The content inconsistency between NIR-RGB images and the scarcity of real-world paired datasets limit its effective application in real scenarios.
In this work, we propose a plug-and-play Selective Fusion Module (SFM) and a Real-world NIR-Assisted Image Denoising (Real-NAID) dataset to address these issues. 
Specifically, SFM sequentially performs global and local modulations on NIR-RGB features before their information fusion.
The Real-NAID dataset is collected with various noise levels under diverse scenes.
Experiments on both synthetic and our real-world datasets show the proposed method achieves better results than state-of-the-art ones.

\section*{Acknowledgments}
  This work was supported by the National Natural Science Foundation of China (NSFC) under Grants No. U22B2035.

\bibliographystyle{IEEEtran}
\bibliography{IEEEabrv, tmm}

\end{document}